\documentclass[pmlr,twocolumn,10pt]{jmlr} 

\usepackage{booktabs}

\DeclareMathOperator{\arctantwo}{arctan2}
\usepackage{gensymb}

\theorembodyfont{\upshape}
\theoremheaderfont{\scshape}
\theorempostheader{:}
\theoremsep{\newline}

\jmlrvolume{LEAVE UNSET}
\jmlryear{2022}
\jmlrsubmitted{LEAVE UNSET}
\jmlrpublished{LEAVE UNSET}
\jmlrworkshop{Conference on Health, Inference, and Learning (CHIL) 2022} 

\title[Using Orientation to Distinguish Overlapping Chromosomes]{Using Orientation to Distinguish Overlapping Chromosomes}

\author{%
\Name{Daniel Kluvanec} \Email{daniel.kluvanec@durham.ac.uk}\\
\Name{Thomas B. Phillips} \Email{thomas.b.phillips@durham.ac.uk}\\
\Name{Kenneth J. W. McCaffrey} \Email{k.j.w.mccaffrey@durham.ac.uk}\\
\Name{Noura Al Moubayed} \Email{noura.al-moubayed@durham.ac.uk}\\
\addr Durham University, United Kingdom
}

\begin{document}

\maketitle

\begin{abstract}
A difficult step in the process of karyotyping is segmenting chromosomes that touch or overlap. In an attempt to automate the process, previous studies turned to Deep Learning methods, with some formulating the task as a semantic segmentation problem. These models treat separate chromosome instances as semantic classes, which we show to be problematic, since it is uncertain which chromosome should be classed as \#1 and \#2. Assigning class labels based on comparison rules, such as the shorter/longer chromosome alleviates, but does not fully resolve the issue. Instead, we separate the chromosome instances in a second stage, predicting the orientation of the chromosomes by the model and use it as one of the key distinguishing factors of the chromosomes. We demonstrate this method to be effective. Furthermore, we introduce a novel Double-Angle representation that a neural network can use to predict the orientation. The representation maps any direction and its reverse to the same point. Lastly, we present a new expanded synthetic dataset, which is based on Pommier's dataset, but addresses its issues with insufficient separation between its training and testing sets.
\end{abstract}

\paragraph*{Data and Code Availability}
This research uses the DeepFish dataset, also known as Overlapping Chromosomes, published by Jean-Patrick Pommier
\footnote{
    Pommier, J.P. (2016) Overlapping chromosomes dataset \\
    \url{https://www.kaggle.com/jeanpat/overlapping-chromosomes} \\
    \url{https://github.com/jeanpat/DeepFISH}
}.

Our code and modified overlapping chromosome dataset is available on GitHub
\footnote{
    Our Code and Dataset repository \\
    \url{https://github.com/KluvaDa/Chromosomes}
}.

\section{Introduction}
\label{sec:introduction}

There are numerous defects that can be identified by visually inspecting chromosomes, such as genetic abnormalities \citep{ChromosomeBook} or cancer \citep{ChromosomeCancer}. Chromosomes are separated and ordered by size from a metaphase image in a process called karyotyping \citep{karyotyping}, which is commonly done by hand and is not yet fully automated. One of the difficulties with karyotyping is that the chromosomes are positioned randomly in the image and can touch or overlap. It is often simpler to prepare multiple images of chromosomes from the same patient and search for ones without overlaps \citep{ChromosomeReview1,shen2019dicentric} than to segment overlapping chromosomes. Numerous image processing algorithms have been proposed to separate overlapping chromosomes \citep{SeparatingChromosomes1,SeparatingChromosomes2,SeparatingChromosomes4,ChromosomeReview2,SeparatingChromosomes3,SeparatingChromosomes5}, however the task remains an open problem. Deep Learning \citep{DeepLearning_LeCun2015}, which has advanced many areas of image processing and computer vision, promises to aid with this task as well.

Semantic segmentation and instance segmentation are two approaches that seem particularly appropriate for the separation of chromosomes. Semantic segmentation classifies every pixel in an image into one of many semantic categories, while instance segmentation additionally classifies multiple instances of the same class separately. We argue that using semantic segmentation to separate overlapping chromosomes is inappropriate, since the chromosome instances are semantically identical. If a category was created for each of the overlapping chromosomes the model would not be able to tell which chromosome should be assigned to which category. We criticise existing methods that chose this approach in \sectionref{sec:related work}. It is possible that assigning a category to the chromosomes based on a rule, such as the longer/shorter chromosome or the more/less vertical chromosome, is sufficient to distinguish the categories. We demonstrate an improvement with such comparison rules, but find that the performance is lacking in comparison to a single semantic category in \sectionref{sec:semantic}.

Highlighting separate instances is done in a variety of ways \citep{instancesegmentation1}, but the methods commonly classify images semantically and use bounding boxes \citep{YOLO} to separate instances. However, bounding boxes rely on different instances being localised differently. This is not necessarily the case for overlapping chromosomes, which may mostly occupy the same space. We argue that a key assumption can be made about chromosomes, which is that in order for two chromosomes to overlap, they must lie in different directions, have different orientations. The only exception to this rule is when two parallel chromosomes touch end-to-end or side-by-side, where knowing their area of intersection is sufficient to segment them. It is for this reason that we explore ways in which a deep learning model can predict the orientation of chromosomes and demonstrate that this information can be used to separate chromosomes in \sectionref{sec:orientation}.

We define the orientation of the chromosome as the direction along its arms, however it does not matter what the top and the bottom of the chromosome is. If a chromosome is rotated  by $180\degree$ we treat it as having the same orientation. This makes it difficult to represent the orientation in a way that a neural network can predict. For this purpose we propose a novel Double-Angle representation in \sectionref{sec:orientation}, which maps any vector and its negative to the same point in space, and train a model to predict it.

We use Pommer's dataset, which consists of 13434 synthetically generated images of overlapping chromosomes, generated from a pool of only 46 chromosomes. This means that the images are not independent, which means that it cannot easily be separated into independent subsets for training and testing. We criticise the dataset in \sectionref{sec:dataset} and papers that used the dataset in this form in \sectionref{sec:related work}. To address the issue, we modify the dataset, reusing Pommier's method to create the synthetic images, but ensure that training and testing subsets don't reuse the same chromosomes. In addition, we annotate the chromosomes with the orientation and expand the set of source images from 46 chromosomes to 620, all of which are based on data provided by Pommier.

In summary, our contributions in this paper are the following:

\begin{itemize}
    \item We present a new bigger synthetic dataset based on Pommier's data, which can be split into independent subsets and contains annotations for the orientation of chromosomes.
    \item We demonstrate that chromosome instances cannot be separated effectively as semantic categories.
    \item We propose a novel representation of orientation called Double-Angle, which maps any vector and its inverse to the same point.
    \item We demonstrate that orientation can be used to separate overlapping chromosomes.
\end{itemize}

\section{Related Work}
\label{sec:related work}

\cite{ChromosomeBaseline1} and \cite{ChromosomeBaseline2} both used Pommier's dataset and a semantic segmentation method. They classify the pixels in the image into four categories: chromosome 1, chromosome 2, overlap and background. The segmentation result would also be valid if the labels chromosome 1 and chromosome 2 were reversed for any given image. Nevertheless, they train the model to predict the specific category that was assigned to every image in Pommier's dataset. \cite{ChromosomeBaseline2} identify that the performance of their system is better if the two chromosomes categories are merged. However, we believe all of the reported metrics to be unreliable because of a design error in Pommier's dataset, which allows superficial performance gains achieved through overfitting to improve performance on the test set. This means that the model would not generalise not only to other datasets, but also to completely unseen images created the same way. We further criticise the dataset in \sectionref{sec:dataset}.

\cite{ChromosomeBaseline3-adversarial} also treated the task as semantic segmentation, however, they addressed the issue of the arbitrary class labels using a cGAN \citep{cGAN} to calculate an additional adversarial loss term using the discriminator network. This loss is agnostic to which class label is assigned to which chromosome, since it is not directly compared against the label.

\cite{ChromosomeBaseline4-ChromSeg} adopted a two-stage method. In the first stage they use a deep learning method to perform a 3-category semantic segmentation with the classes background, chromosome and overlap. In the second stage they separate the chromosomes instances using an image processing algorithm. On top of segmenting where the individual chromosomes are, \cite{ChromosomeBaseline4-ChromSeg} also reconstructs an image of the separated chromosomes. 

\section{Dataset}
\label{sec:dataset}
Pommier's dataset is created from a single image of a human metaphase stained using 4',6-diamidino-2-phenylindole together with a Cy3 fluorescent telo\-meric probe \citep{cy3}. The image contains 46 individual chromosomes that do not overlap or touch. The synthetic dataset is created by taking pairs of chromosomes, rotating, translating, and averaging them. The synthetic images are then labelled with the categories: chromosome 1, chromosome 2, overlap and background, where any pair of chromosomes will always have the same labels regardless of their rotation and translation. The model could therefore recognise a pair of chromosomes in the test set and recall their class labels, having seen them before, albeit rotated and translated differently. In other words, the dataset cannot be split into independent subsets, which would allow overfitting on one subset to benefit the performance on the other. As with all models trained on synthetic datasets, the extent to which they generalise to other datasets is questionable, however, when using Pommier's training and testing sets, even the extent to which the model generalises to the synthetic images is uncertain.

To address this issue, we propose that the chromosome source images are separated into subsets, before selecting pairs to create synthetic images with. Pommier also published 14 other human metaphase images that were not used to create the dataset, some of which contain overlapping chromosome pairs and larger clusters. We manually segmented all 15 metaphase images, which contained 620 separate chromosomes, while also annotating their orientation. In our experiments, 20\% of the data is reserved for testing, while the remaining 80\% is split into training and validation sets using 4-fold cross validation \citep{crossvalidation}. The synthetic images are created in a similar way to Pommier's method.

The methaphase images also contained 30 images of overlapping chromosome pairs and larger clusters. We use these to tell how well the model generalises from synthetic to real data, while minimising the domain shift that would occur when comparing against other datasets, which we use in equal proportion for validation and testing. Our orientation-based segmentation can work effectively with clusters, however the semantic segmentation can only separate pairs of chromosomes. To test the semantic segmentation, we separated the larger clusters into a total of 41 smaller images that are cropped around each pair of chromosomes, expecting the model to only classify the chromosome pair and ignore the other chromosomes that might be visible in the image. 

\section{Model}
\begin{figure*}[t]
\floatconts
{fig:unet}
{\caption{The U-net-style neural network architecture used, where the numbers in the image represent the number of channels.}}
{\includegraphics[width=0.8\textwidth]{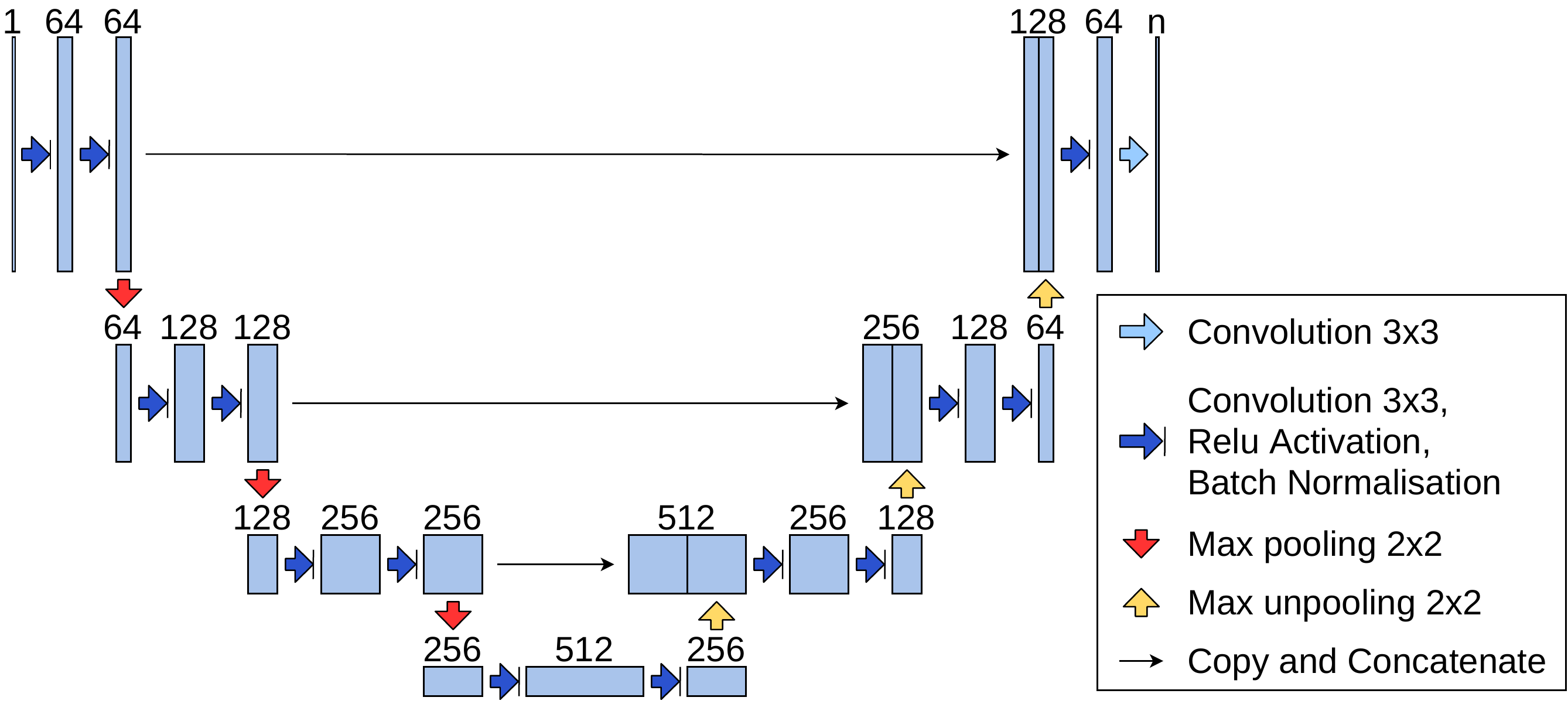}}
\end{figure*}

\label{sec:model}
We ran our experiments with both U-net \citep{unet} network variations used by \cite{ChromosomeBaseline1} and by \cite{ChromosomeBaseline2}. We found that the larger network performs marginally better and report on its results. To adapt the architecture to the different task definitions, we only modify the number of output channels and the loss function used. The network architecture is illustrated in Figure \ref{fig:unet}.

We use 4-fold cross validation \citep{crossvalidation}, where we train the model independently 4 times, using a different quarter of the training set for validation. All metrics are averaged across the 4 runs.

We train the model using an Adam optimizer \citep{adam}, with a learning rate of 0.001 and betas of 0.9 and 0.999.

\section{Semantic Segmentation}

\begin{table*}[t]
\caption{Semantic segmentation results, shown as IOU \% scores. The individual semantic categories are: background, chromosome 1, chromosome 2 and overlap. ch1+ch2 represents the area covered by merging the ch1 and ch2 categories. }
\centering
\begin{tabular}{p{2cm} p{2cm} p{1.5cm} p{0.9cm} p{0.9cm} p{0.9cm} p{1.5cm} l}
\toprule[1pt]

\bfseries Training & \bfseries Testing & \bfseries Average & \multicolumn{4}{l}{\bfseries Individual IOUs} & \bfseries ch1+ch2 \\
\bfseries Dataset  & \bfseries Dataset & \bfseries IOU     & back. & ch1 & ch2 & over.    & {}         \\

\midrule[1pt]
Pommier’s       & Pommier’s & \bfseries 92.4      & 100   & 90.7  & 96.4  & 82.6  & 97.7  \\
                & Synthetic & 43.2      & 86.6  & 26.1  & 25.2  & 34.9  & 44.4  \\
                & Real      & 35.9      & 66.5  & 25.8  & 22.3  & 29.0  & 51.4  \\
\hline
Synthetic       & Pommier’s & 51.1      & 92.4  & 33.1  & 48.3  & 30.5  & 55.2  \\
Length-         & Synthetic & 71.2      & 98.8  & 69.5  & 56.9  & 59.8  & 88.1  \\
wise            & Real      & 50.2      & 84.3  & 42.1  & 38.7  & 35.8  & 67.3  \\
\hline
Synthetic       & Pommier’s & 60.0      & 97.3  & 44.8  & 58.8  & 39.2  & 63.5  \\
Orientation-    & Synthetic & 75.3      & 98.8  & 72.2  & 70.5  & 59.7  & 87.7  \\
wise            & Real      & \bfseries 56.8      & 84.2  & 49.8  & 53.4  & 39.8  & 67.7  \\
\hline
Synthetic       & Pommier’s & 55.2      & 97.5  & 38.7  & 47.9  & 36.6  & 62.2  \\
Position-       & Synthetic & \bfseries 77.3      & 98.8  & 74.2  & 76.2  & 60.0  & 88.0  \\
wise            & Real      & 52.7      & 84.5  & 43.9  & 43.3  & 39.2  & 67.9  \\
\hline
Synthetic       & Pommier’s & 50.9      & 95.5  & 29.5  & 46.3  & 32.1  & 61.3  \\
Random          & Synthetic & 64.6      & 98.7  & 53.4  & 52.1  & 54.0  & 87.5  \\
                & Real      & 46.5      & 82.3  & 35.6  & 34.9  & 33.5  & 65.1  \\
\bottomrule[1pt]
\end{tabular}
\label{tab: semantic segmentation}
\end{table*}

\begin{figure*}[t]
\floatconts
{fig:semantic segmentation}
{\caption{Semantic segmentation result examples. The semantic classes are visualised as: blue - background, green/orange - chromosome 1/chromosome 2, red - overlap.}}
{\includegraphics[width=0.85\textwidth]{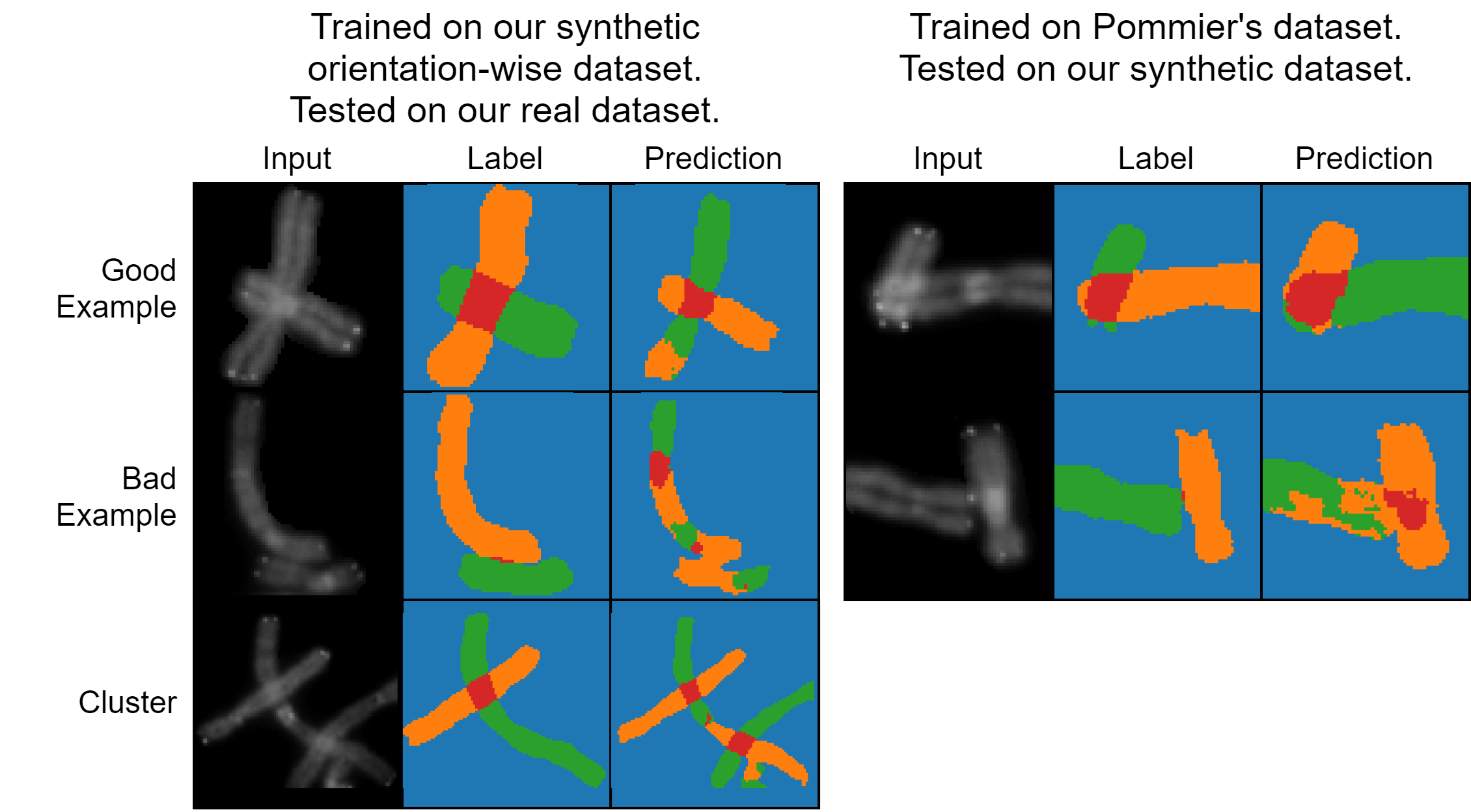}}
\end{figure*}

\label{sec:semantic}
We argue that separating chromosome instances using a semantic segmentation with the classes: background, chromosome 1, chromosome 2 and overlap is an inappropriate approach, because the chromosome instances aren't semantically different. This is problematic because it is uncertain which chromosomes should be assigned which class label. We hypothesise that this issue could be addressed by applying comparison rules for determining the class label, even though the chromosomes remain semantically similar. We tested the following properties for comparing the chromosomes:

\begin{itemize}
    \item length-wise, as the longer/shorter chromosome
    \item orientation-wise, as the more/less vertical chromosome
    \item position-wise, as the rightmost/leftmost chromosome
    \item randomly as a control
\end{itemize}

Despite using these comparison rules to define which class label to assign during training, it does not matter if the labels were switched during evaluation. We therefore switch the predicted classes for chromosome 1 and chromosome 2 if it yields better metrics. The metrics that we track are the intersection over union (IOU) scores, also known as the Jaccard Index \citep{jaccard}. These are calculated for every predicted class separately and averaged to get an overall performance metric. For comparison with our orientation-based segmentation, we also report the IOU scores averaged over the two chromosome categories, chromosome 1 and chromosome 2, only.

Table \ref{tab: semantic segmentation} presents the results of the semantic segmentation. Example images of good and bad results can be seen in \figureref{fig:semantic segmentation}. The following are our key observations:

\begin{itemize}
    \item When training and evaluating on Pommier's dataset, we achieve a semantic IOU score of 92.4\%, which is comparable to 94.7\% reported by \cite{ChromosomeBaseline1} and the range of 90.63\% - 99.94\% reported by \cite{ChromosomeBaseline2}.
    \item There remains a sizeable domain shift between Pommier's synthetic dataset and ours despite their similar approach. Transitioning from Pommier's to our synthetic dataset yields a 49.2\% reduction in average IOU, while the opposite yields a reduction of at least 13.7\%.
    \item The average IOU score when trained and evaluated on our synthetic dataset is at least 15.1\% lower than when trained and evaluated on Pommier's dataset. This suggests that either our synthetic dataset is more difficult to solve, or that the performance metrics on Pommier's dataset are unreliable due to overfitting improving testing performance.
    \item To compare the rules for assigning class labels, we look at their average IOU on the synthetic and real datasets. Orientation-wise assignment yielded the best performance on the real dataset, while position-wise assignment yielded the best performance on the synthetic dataset. Notably, all three rules performed better than the control with random assignment, with an improvement of 6.6\%, 10.7\% and 12.7\% for the synthetic dataset and an improvement of 3.7\%, 10.3\% and 6.2\% for the real dataset, respectively for length-wise, orientation-wise and position-wise assignment.
    \item Another way of comparing how well the assignment rules work is to compare the IOU score for the two chromosome classes with the IOU score if the categories were merged. For the synthetic dataset we find a difference of 16.6\%, 9.4\%, 6.2\% and 25.2\%, and for the real data 20.3\%, 10.7\%, 17.6\% and 23.4\%, respectively for length-wise, orientation-wise, position-wise and random assignment. Larger values suggest that the model is more confused with which of the two chromosome labels to assign. These results similarly show that position-wise assignment performed best on the synthetic dataset, while orientation-wise assignment performed best on the real dataset.
\end{itemize}

Despite the improvement that the class assignment rules offer, we believe that the approach is still fundamentally problematic. Firstly, there are instances where the properties used for the assignment rules are very similar for both chromosomes. Two chromosomes can have the same length, orientation or position. In these cases the assignment is still a source of uncertainty. Secondly, using a semantic segmentation in this way requires the image to be cropped around the pair of overlapping chromosomes. This could be done with a separate model, however it also causes the issues with larger clusters, where more than two chromosomes are visible in the cropped image, which the model wasn't trained to deal with. This can also be seen in \figureref{fig:semantic segmentation}, where the model is expected to only highlight two chromosomes in the cluster, but struggles to tell which chromosomes cropped around. We would therefore recommend only using the semantic segmentation to distinguish the semantic categories: background, unique-chromosome and overlap, which can be applied to images with any amount of chromosomes, and rely on other methods to separate chromosome instances.

\section{Orientation-Based Segmentation}
\label{sec:orientation}
\begin{figure*}[t]
\floatconts
{fig:instance segmentation}
{\caption{Orientation-based segmentation result examples. The Labels and Predictions are the separated chromosomes, while Semantic, Dilated Overlap and Orientation are the network outputs. The semantic classes are represented as: blue - background, orange - chromosome, green - overlap. The Orientation is colour-coded.}}
{\includegraphics[width=0.85\textwidth]{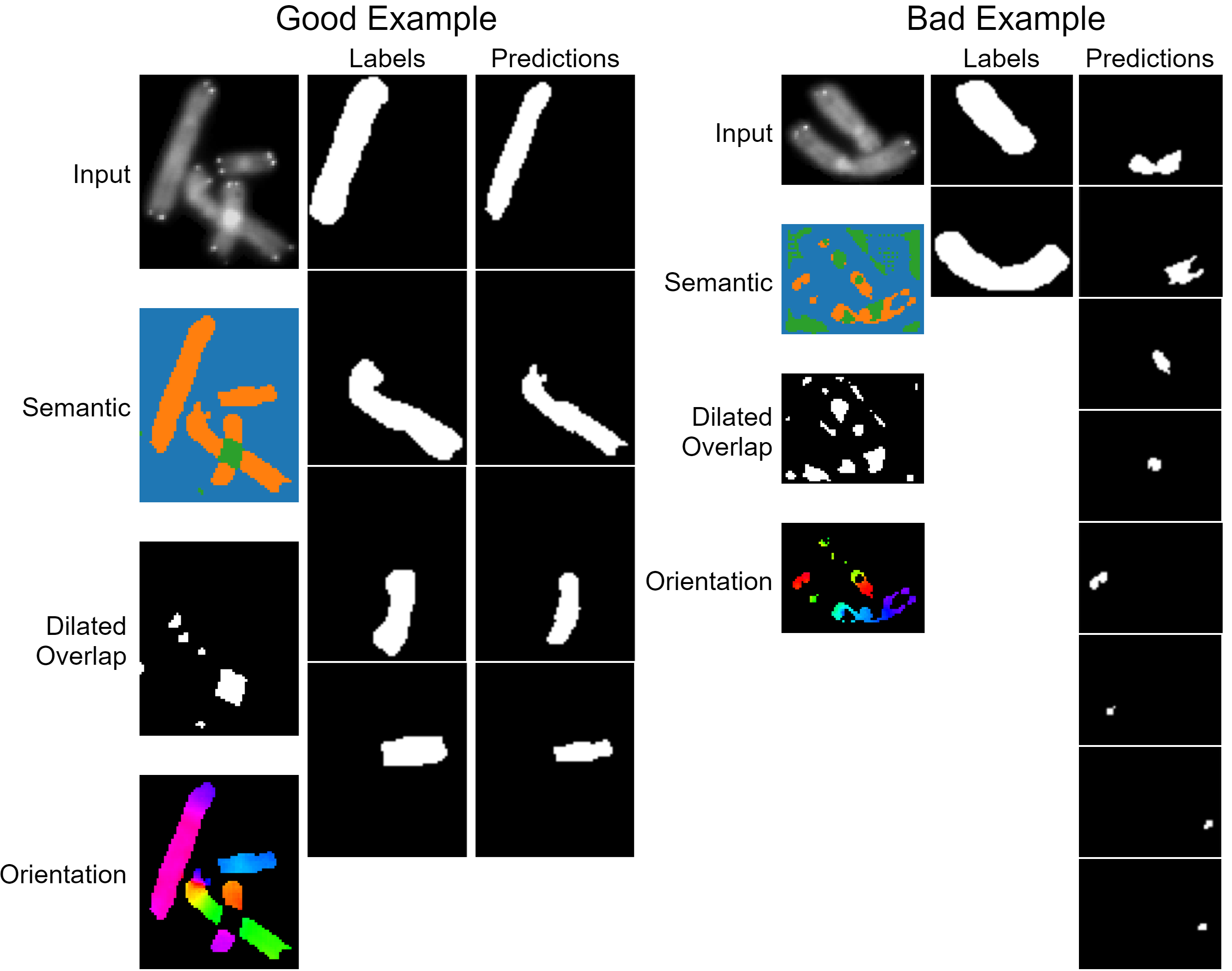}}
\end{figure*}

In order to segment chromosome instances, we argue that knowing their orientation is an important distinguishing feature, since most overlapping chromosomes will have different orientations. The only other case being parallel chromosomes that partially overlap, either side-to-side or end-to-end. In this case the overlapping area should fully separate the two chromosome instances. The key utility provided by knowing the orientation of the chromosomes is, that it can be used to match multiple separate chromosome areas that are divided by overlapping areas. Assuming that chromosomes don't bend very sharply, the orientation of the chromosome before and after an overlap will be mostly the same.

We define the orientation of the chromosome as a direction vector along the length of the chromosome. However it does not matter what the top and the bottom of the chromosome is. If the chromosome is rotated by 180\degree it will still have the same orientation. Equivalently, any direction vector and its negative represent the same orientation. Equivalently, the orientation could be represented as an angle in the range $[0\degree, 180\degree)$. Getting a neural network to predict such an orientation is non-trivial, which led us to develop a novel Double-Angle representation.

Getting a neural network to predict the value of an angle directly is difficult, because of its cyclical nature and a discontinuity at $0\degree$ and $360\degree$. Angles close to the discontinuity are very similar, yet their angle values can be very different. Instead one can represent an angle $\theta$ using $\sin(\theta)$ and $\cos(\theta)$, which is effectively a unit vector positioned $\theta$ degrees anticlockwise from the x-axis. This definition avoids the issues of discontinuity. In our case, with angles restricted to $180\degree$, the resulting representation would effectively be a unit vector with a positive y-axis component, which would still contain a discontinuity near values of $(1, 0)$ and $(-1, 0)$. Instead let us define the Double-Angle representation that maps any vector and its negative onto the same point as follows:
\begin{align}
\begin{split}
    \texttt{Double-Angle}\colon \: & \\
    [0\degree, 180\degree) \longrightarrow & \mathbb{R}^2\\
    \theta                 \longmapsto & (\cos(2\theta), \sin(2\theta)),
    \end{split} \\
\intertext{with its inverse:}
\begin{split}
    \texttt{Double-Angle}^{-1}\colon \: & \\
    \mathbb{R}^2 \longrightarrow & [0\degree, 180\degree) \\
    (d_0, d_1)   \longmapsto & \frac{1}{2}\arctantwo(d_1, d_0) \mod{180\degree}.
\end{split}
\end{align}

Our proposed instance segmentation model outputs a total of six channels. Two channels are used for the Double-Angle representation, which is trained using the Huber loss \citep{huberloss} and masked out in pixels that do not belong to a unique chromosome. The third channel is used to predict a dilated overlap, which is defined as the overlapping area of the chromosomes dilated by two pixels. This allows us to separate chromosomes that are very close to each other but do not overlap. It is trained using a binary cross-entropy loss (with logits). The last three channels are used as a semantic segmentation with the classes: background, chromosome and overlap, which is trained using a cross-entropy loss. These predictions are subsequently used by our instance segmentation post-processing algorithm, which uses image-processing to separate the chromosome instances. Its intended purpose is to demonstrate the utility of the Double-Angle representation and its use for segmenting chromosome instances. The algorithm is defined in Algorithm \ref{algorithm} and explained below:

\begin{algorithm2e*}[t]
\label{algorithm}
\caption{Orientation-Based Instance Segmentation}
\KwIn{\textit{orientation}, \textit{dilated\_overlap}, \textit{background}, \textit{chromosome}, \textit{overlap}}
remove small areas from \textit{overlap}\\
\textit{distance\_image} $\leftarrow$ distance transform of (\textit{chromosome} - \textit{dilated\_overlap})\\
\textit{segments} $\leftarrow$ local maxima of (\textit{distance\_image})\\
\For{\upshape \textit{distance} in max(\textit{distance\_image}) to 0}{
    dilate \textit{segments} until they fill the area where \textit{distance\_image} $\geq$ \textit{distance}\\
    \If{\upshape
    two \textit{segments} touch and maximum distance in either \textit{segment} $\leq$ \textit{distance}$+ 1$
    }{
        merge the two \textit{segments}\\
    }
}
dilate \textit{segments} until they fill the area of \textit{chromosome}\\
\For{\upshape all pairs of \textit{segments}}{
    \textit{neighbouring\_pixels} $\leftarrow$ intersection of (pair of \textit{segments} dilated by 1 pixel)\\
    \If{\upshape
    \textit{neighbouring\_pixels} are not near an overlap and their \textit{orientation} is sufficiently similar
    }{
        merge the two \textit{segments}
    }
}
remove small \textit{segments}\\
\For{\upshape all \textit{overlaps}}{
    \While{\upshape there are more than two \textit{segments} adjacent to the \textit{overlap}}{
        merge the \textit{segments} with the most similar \textit{orientation}
    }
}

\textit{separate\_chromosomes} $\leftarrow$ \textit{segments} merged with adjacent areas of \textit{overlap}
\end{algorithm2e*}

In order to separate the chromosome instances we first analyse the prediction and identify unique chromosome segments. These segments are then merged to create the predicted chromosome instances based on their orientation and relative location to the areas of overlap.

To avoid issues with noisy network outputs, we remove any areas of overlap that are too small. We then subtract the dilated overlap prediction from the chromosome class to spatially separate chromosome segments from each other. Nonetheless, the segments may not be fully disjointed, so we perform the following operation to identify them. First we determine seeding points as the maxima of a distance transform applied to the segments. These will be the points in which the chromosome segments are the widest. We then iteratively grow these segments to cover the whole segment area. Crucially, due to fluctuations in chromosome widths, multiple seeding points may exist in a single chromosome segment. We therefore merge two segments if either of their seeding points was barely a local maximum, raised only 1 pixel over the neighbouring area. Finally we grow the segments over the areas of the dilated overlap until they cover the whole chromosome areas. We further consider merging two adjacent segments if their orientation is sufficiently similar in the area where they touch. We do not apply this operation near areas of overlap where two distinct chromosomes must be present. Next we determine which chromosome segments belong to the same chromosome. We assume that every area of overlap is created by overlapping exactly two chromosomes. We therefore merge the chromosome segments with the most similar orientation until only two chromosome instances are present near every overlap. Finally we add the adjacent overlap areas to all chromosome instances.

This algorithm uses the output of a model that achieved the performance metrics seen in \tableref{tab:instance model}. After separating the chromosomes using the instance segmentation algorithm, we compare the labelled chromosomes with the best matching predicted chromosome and average their IOU values. Note that any extra predicted chromosomes that did not get matched are ignored. These metrics can be seen in \tableref{tab:instance result} together with comparable results from the semantic segmentation model trained on the orientation-wise synthetic dataset. The instance segmentation performs 7.1\% better on the synthetic data and 20.8\% better on the real data than the semantic segmentation. When qualitatively evaluating the results in \figureref{fig:instance segmentation}, we find that the model can correctly separate some examples, but fails on others. When it does fail, the network prediction is very noisy, suggesting that the model failed to generalise well. We suspect that the issue stems from the synthetic dataset, which does not perfectly resemble real chromosome images. Moreover, the dataset is of limited size and sample variety, owing to its source of only 15 human metaphase images, nine of which were used for training.

\begin{table}[ht]
\floatconts
  {tab:instance model}
  {\caption{The performance metrics achieved by the model used for orientation-based instance segmentation}}
  {\begin{tabular}{ll}
  \toprule
  \bfseries Outputs & \bfseries Metric Value\\
  \midrule
  Background IOU & 98.8\%\\
  Chromosome IOU & 88.3\%\\
  Overlap IOU & 58.6\%\\
  Dilated Overlap IOU & 70.4\%\\
  Orientation Difference & 5.2\degree\\
  \bottomrule
  \end{tabular}}
\end{table}

\begin{table*}[t]
  \centering
  \caption{The average IOU \% scores over the best-matching separated chromosomes.}
  \label{tab:instance result}
  \begin{tabular}{l l l}
  \toprule
  \bfseries Testing & \bfseries Semantic Segmentation  & \bfseries Orientation-Based  \\
  \bfseries Dataset & \bfseries (orientation-wise) & \bfseries Segmentation \\
  \midrule
  Pommier's & 61.5 & 66.4 \\
  Synthetic & 78.3 & 85.4 \\
  Real & 57.0 & 77.8 \\
  \bottomrule
  \end{tabular}
\end{table*}

\section{Conclusion}
\label{sec:conclusion}
In this paper we address the task of segmenting overlapping chromosomes. When using an existing synthetic dataset, we warn of its images not being independent and therefore not appropriate for splitting into training and testing subsets. We present an alternative synthetic dataset based on the same source images that has proper separation between training, validation and testing sets.

We demonstrate that segmenting chromosome instances as separate semantic classes is problematic. This is especially true when class labels for chromosome 1 and chromosome 2 are randomly assigned. Applying comparison rules to assign the class improves performance, but remains problematic in cases where the overlapping chromosomes are too similar. Moreover this method only works well for pairs of overlapping chromosomes and not clusters.

Instead, we propose separating the chromosome instances out in a post-processing step, which relies on additional predicted features from the deep learning model. Most importantly, we rely on the orientation of the chromosomes to separate them out. We propose a novel Double-Angle representation, which a neural network can use to predict the orientation of the chromosomes. The Double-Angle representation maps any direction vector in 2D and its negative onto the same point in a continuous and smooth manner. We demonstrate the effectiveness of separating the chromosomes based on their orientation and encourage readers to incorporate orientation as one of the distinguishing factors in their approaches.

For future work we suggest focusing on improving the deep learning model's performance by improving the data used to train the model. Larger datasets that include real images of overlapping chromosomes in the training sets could make the model generalise better. Data augmentation, which would add noise and distortions to the images, could also serve to make the model more robust.

\section*{Institutional Review Board (IRB)}
This research does not require IRB approval.

\acks{This work was supported by Durham University, the European Regional Development Fund–Intensive Industrial Innovation Programme Grant No. 25R17P01847 and GeoTeric Ltd.}

\bibliography{bibliography}

\end{document}